\documentclass[letterpaper, 10 pt, conference]{ieeeconf}  
\usepackage[font=footnotesize,labelfont=bf]{caption}

\IEEEoverridecommandlockouts                              

\usepackage{graphics} 
\usepackage{epsfig} 
\usepackage{mathptmx} 
\usepackage{times} 
\usepackage{amsmath} 
\usepackage{amssymb}  
\usepackage{booktabs}       
\usepackage{afterpage}
\usepackage[table]{xcolor}
\usepackage{afterpage}
\usepackage[dvipsnames]{xcolor} 
\usepackage{multirow}
\usepackage{multicol}
\usepackage{pifont} 
\usepackage{placeins}
\usepackage{caption}
\usepackage{cite}
\usepackage{stfloats} 
\usepackage{titlesec}
\usepackage{hyperref}

\title{\LARGE \bf
A Self-Supervised Approach on Motion Calibration\\for Enhancing Physical Plausibility in Text-to-Motion
}

\author{Gahyeon Shim$^{1}$, Soogeun Park$^{1}$, and Hyemin Ahn$^{2}$
\thanks{$^{1}$Gahyeon Shim and Soogeun Park are with Artificial Intelligence Graduate School (AIGS),
Ulsan National Institute of Science and Technology (UNIST), Ulsan, Korea
        (email: {\tt\small gahyeon@unist.ac.kr, soogeun@unist.ac.kr})}%
\thanks{$^{2}$Hyemin Ahn is with Department of Electrical Engineering (EE), 
        Pohang University of Science and Technology (POSTECH), Pohang, Korea
        (email: {\tt\small hmahn@postech.ac.kr)})}%
}

\begin{document}

\maketitle
\thispagestyle{empty}
\pagestyle{empty}


\begin{abstract}
Generating semantically aligned human motion from textual descriptions has made rapid progress, but ensuring both semantic and physical realism in motion remains a challenge. In this paper, we introduce the \textbf{Distortion-aware Motion Calibrator (DMC)}, a post-hoc module that refines physically implausible motions (e.g., foot floating) while preserving semantic consistency with the original textual description. Rather than relying on complex physical modeling, we propose a self-supervised and data-driven approach, whereby DMC learns to obtain physically plausible motions when an intentionally distorted motion and the original textual descriptions are given as inputs. We evaluate DMC as a post-hoc module to improve motions obtained from various text-to-motion generation models and demonstrate its effectiveness in improving physical plausibility while enhancing semantic consistency. The experimental results show that DMC reduces FID score by 42.74\% on T2M and 13.20\% on T2M-GPT, while also achieving the highest R-Precision. When applied to high-quality models like MoMask, DMC improves the physical plausibility of motions by reducing penetration by 33.0\% as well as adjusting floating artifacts closer to the ground-truth reference. These results highlight that DMC can serve as a promising post-hoc motion refinement framework for any kind of text-to-motion models by incorporating textual semantics and physical plausibility. 
Project page: \url{https://sites.google.com/view/motion-calibrator/}
\end{abstract}

\section{INTRODUCTION}

 
The rapid advancement of deep generative models has led to remarkable progress in human motion generation. Conditional human motion generation, which synthesizes motion from high-level inputs such as textual descriptions, enables the creation of diverse and context-aware motions beyond simple, repetitive patterns. In the domain of animation, this approach streamlines avatar creation through natural language customization, reducing both production costs and required expertise~\cite{hong2022avatar, makeananimation2023, Sketch2Anim2025}. Within virtual reality, it further enhances immersion by supporting virtual agents that can produce contextually appropriate gestures and actions directly from text~\cite{bha2021vr, toward2024}.

In robotics, the potential impact is particularly significant. By learning from human motion data, robots can acquire loco-manipulation skills that integrate mobility and object interaction in a more intuitive and human-like manner~\cite{ imitationnet2023, jiang2024harmon, he2024omnih2o, gu2025humanoidlocomotionmanipulationcurrent}. This capability allows even non-expert users to control robots through natural language commands, lowering the barrier to everyday deployment. Furthermore, such models lay the groundwork for natural and seamless human–robot interaction, which is essential for collaborative and socially aware robotic systems.

However, existing motion generators often fail to synthesize physically plausible motions. Although their motions appear semantically appropriate when language is provided, they frequently contain physical artifacts, such as foot skating, floating, clipping, and ground penetration. These artifacts are caused by overly smooth transitions between poses, incorrect foot-ground contact, or unnatural foot-to-foot interactions. These physically implausible patterns severely limit the usability of generated motions in real-world scenarios, especially in physically interactive or grounded applications. In the case of humanoid robots, retargeting motions that contain physical artifacts can lead to unstable behaviors, excessive mechanical load, or even safety hazards. Consequently, achieving physically plausible motion is not only a matter of realism but also a prerequisite for extending the applicability of motion generators to diverse and practical domains.
 
To mitigate the issue, recent works explored ways to enhance the physical plausibility of generated motions. One line of work~\cite{tevet2023human} introduces auxiliary loss functions (i.e., foot-to-plane distance loss) to encourage physical alignment, while others~\cite{yuan2023physdiff, Han_2025_WACV, kang2025biomodiffuse} incorporate physical constraints directly into the generative process through reinforcement learning, reward-guided refinement, or physics guidance. There also exist methods~\cite{tashakori2025flexmotionlightweightphysicsawarecontrollable} which leverage an unsupervised physics discriminator that implicitly learns physical plausibility through motion cues such as contact timing and joint dynamics.
While these improved physical realism, they depend on costly simulations and complex reward design~\cite{yuan2023physdiff, Han_2025_WACV}\, or on simplified heuristics that limit generalizability~\cite{tashakori2025flexmotionlightweightphysicsawarecontrollable}. As a result, ensuring fine-grained physical plausibility in a scalable and efficient manner remains an open challenge.

To address this gap, we propose a \textbf{post-hoc refinement framework} called \textbf{Distortion-aware Motion Calibrator (DMC)}, which is for text-to-motion models. Rather than modifying the generative model itself or relying on complicated physics modeling, DMC takes the motions generated by any text-to-motion models and corrects physically implausible artifacts in a learning-based way, while preserving the original semantic intent and stylistic expressiveness described by the original text description. This makes our framework retain the effect of the original generation model while significantly enhancing physical realism, making it both practical and extensible.

DMC is trained in a self-supervised and data-driven manner by applying synthetic distortions to high-quality motions from the HumanML3D dataset~\cite{Guo_2022_CVPR}, creating physically implausible motion samples for self-supervision.
We employ two main distortions: (1) biased ground offsets, which vertically shift the character’s position, resulting in floating or ground penetration artifacts, and (2) temporal smoothing, which removes high-frequency motion details and causes foot sliding. We further introduce two model variants of DMC to accommodate different application needs: (1) a WGAN-based DMC, which offers faster refinement, and (2) a denoising-based DMC, which provides finer-grained correction at the cost of slower inference due to its step-wise refinement process. This dual design enables the selection of the appropriate model variant based on the level of physical artifacts or the demands of the target application.

Notably, DMC provides consistent improvements across text-to-motion baselines with varying levels of motion quality, ranging from artifact-prone models such as T2M~\cite{guo2022generating} and T2M-GPT~\cite{zhang2023t2mgptgeneratinghumanmotion} to high-quality models like MoMask~\cite{guo2023momask}.
The WGAN-based DMC achieves a 42.74\% reduction in FID on T2M and improves R-Precision on the same baseline, significantly enhancing perceptual quality and semantic alignment.
The denoising-based DMC further improves FID on T2M-GPT by 13.20\%, while also excelling at correcting physical artifacts, reducing ground penetration by 42.57\% on T2M, 10.84\% on T2M-GPT, and 33.0\% on MoMask.
It also adjusts floating values closer to the ground-truth reference, indicating improved ground contact. These results demonstrate that DMC effectively refines physically implausible motions without compromising semantic consistency, and is robustly applicable to diverse text-to-motion generation models.
The main contributions of our DMC are as follows:
\begin{itemize}
    \item Distortion-aware Motion Calibrator (DMC) runs without explicit physics modeling, yet preserves the original motion's expressive capacity and semantic fidelity.
    \item DMC is lightweight and model-agnostic, which can be seamlessly integrated with various text-to-motion models.
    \item We introduce two variants of DMC: (1) a WGAN-based one that excels in enhancing perceptual quality and semantic consistency, and (2) a denoising-based one that provides more fine-grained correction of physical artifacts.
    \item Our experiment result shows that DMC can consistently improve the physical plausibility of generated motions from various baseline text-to-motion models without degrading semantic alignment.
\end{itemize}

\afterpage{
\begin{figure*}[t]
  \centering
  \includegraphics[width=0.85\linewidth]{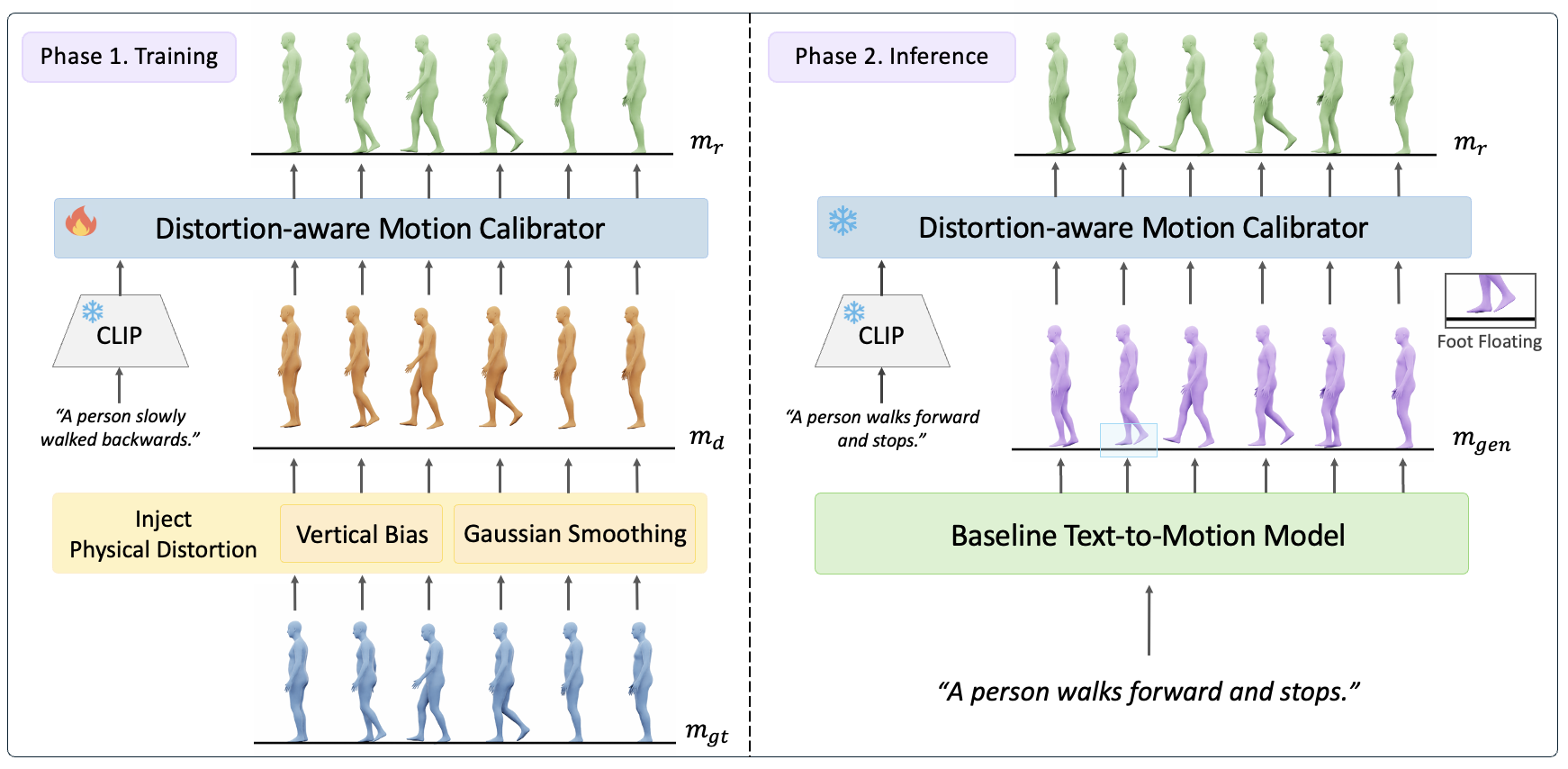}
  \caption{Two phases of how the proposed \textbf{Distortion-aware Motion Calibrator (DMC) works}. \textbf{(1) Training Phase (left)}: The physically plausible ground-truth motions ($\mathbf{m}_{\text{gt}}$) are synthetically distorted via vertical bias and smoothing to produce artifact-laden motions ($\mathbf{m}_{\text{d}}$). DMC learns to refine $\mathbf{m}_{\text{d}}$ into physically plausible outputs ($\mathbf{m}_{\text{r}}$), conditioned on the original textual description. \textbf{(2) Inference Phase (right)}: DMC is applied as a post-hoc module to correct artifacts in motions ($\mathbf{m}_{\text{gen}}$) from any pretrained text-to-motion model, yielding the refined motions ($\mathbf{m}_{\text{r}}$). The framework is lightweight, model-agnostic, and requires no modification to the original generation pipeline.}
  \label{fig:overview}
  \vspace{-2mm}
\end{figure*}
}

\section{RELATED WORK}

\subsection{Human Motion Generation}

Early studies of human motion generation focused on deterministic methods such as physics-based simulation~\cite{badler1993simulating} and keyframe animation~\cite{character2016}. With the rise of deep learning, sequence modeling techniques using RNN~\cite{Fragkiadaki_2015_ICCV}, GCN~\cite{Mao_2019_ICCV}, and MLP~\cite{Bouazizi_2022_IJCAI} became prevalent, offering structurally stable outputs but often producing blurry and monotonous motions due to their inability to capture the stochastic nature of human behavior. To address this limitation, probabilistic generative models like GAN~\cite{cai2018deep}, VAE~\cite{zhang2023remodiffuse}, and diffusion models~\cite{zhang2022motiondiffuse} were introduced, enabling diverse and realistic synthesis by learning complex temporal distributions.


Subsequent research emphasized conditioned motion generation, where models respond to external signals for greater controllability and semantic grounding. Conditioning modalities include pose sequences~\cite{liu2022investigating}, images~\cite{chen2022learning}, action classes~\cite{wang2020learning}, audio~\cite{gong2023tm2d}, and especially textual description~\cite{guo2022generating, zhang2023t2mgptgeneratinghumanmotion, guo2023momask}, valued for its flexibility in describing motion semantics. Early text-to-motion models used RNN-based encoder-decoder frameworks~\cite{ahn2018text2action}, while later approaches adopted GANs~\cite{gong2023tm2d}, VAEs~\cite{petrovich2022temos}, Transformers~\cite{zhang2023motiongpt, guo2023momask}, and diffusion models~\cite{tevet2023human}, which excel at capturing fine-grained temporal dynamics.

Despite these advances, current text-to-motion models still suffer from physically implausible artifacts, such as foot sliding, floating, clipping, and ground penetration, especially when the generated motion deviates from natural contact dynamics. This highlights the necessity of flexible solutions to enhance physical plausibility without modifying the pretrained generative models.

\titlespacing*{\subsection}{0pt}{0.5ex plus 0.1ex minus 0.1ex}{0.5ex}

\subsection{Physical Plausibility in Motion Generation}

To generate physically plausible motion, early research combined reinforcement learning (RL) with physics-based simulation. DeepMimic~\cite{peng2018deepmimic}, SFV~\cite{peng2018sfv}, and related approaches~\cite{liu2017learning, liu2018learning, merel2017learning} demonstrated that RL agents could imitate motion capture data across diverse behaviors, demonstrating that realistic motion can be synthesized using simulation alone. Building on this foundation, later works~\cite{bergamin2019drecon, tevet2025closd, Zhao:DartControl:2025, park2019learning, won2020scalable} extended RL-based frameworks toward user-controllable motion, adapting policies to respond to text, trajectories, or stylistic variations.


Recent efforts integrate physical modeling into generative processes. PhysDiff~\cite{yuan2023physdiff} incorporates a physics engine during inference to guide denoising with Newtonian-consistent corrections, achieving high accuracy but at significant computational cost and requiring a simulator. ReinDiffuse~\cite{Han_2025_WACV} instead optimizes physically inspired reward functions, effectively reducing artifacts such as foot sliding, though it depends on manually designed rewards. BioMoDiffuse~\cite{kang2025biomodiffuse} advances fidelity by embedding biomechanical constraints into diffusion. While this design enables dynamically consistent motion, it increases model complexity and depends on biomechanical data that are often unavailable.


Another line of work focuses on post-hoc refinement to improve temporal stability and perceptual quality without strictly enforcing physics. SmoothNet~\cite{zeng2022smoothnet} applies temporal filtering, while Skeletor~\cite{jiang2021skeletor} and FlexMotion~\cite{tashakori2025flexmotionlightweightphysicsawarecontrollable} use skeletal consistency and attention to correct artifacts. FlexMotion further employs a self-supervised physics discriminator, enabling lightweight improvements but without guaranteeing globally consistent dynamics. These methods provide lightweight perceptual gains but lack physical grounding and generalize poorly to complex motions.


These approaches illustrate trade-offs between physical fidelity, computational cost, and complexity. To address this challenge, we propose a lightweight post-hoc framework that enhances physical plausibility of pretrained text-to-motion models while preserving their generative expressiveness.

\section{METHOD}
Distortion-aware Motion Calibrator (DMC) is based on self-supervised learning, and is trained to correct emulated artifacts that are commonly obtained from modern text-to-motion models. It restores physically plausible motion while preserving semantic alignment with the given original textual description. Once being trained, DMC can be flexibly integrated into any pretrained text-to-motion models.
Figure~\ref{fig:overview} illustrates an overview of the DMC framework. It consists of two phases: (1) a self-supervised training phase, where it learns to recover physically plausible motions from synthetically distorted motions, and (2) an inference phase, where DMC refines the output motion of any kind of text-to-motion model. In both cases, DMC jointly conditions on motion and the corresponding textual description to ensure semantic consistency while improving physical plausibility.

\subsection{Model Architecture}
Let $\mathbf{m}_{d} \in \mathbb{R}^{T \times d_p}$ denote a distorted motion sequence, where $T$ is the number of frames and $d_p$ is the dimension of the pose feature in each frame. In addition, let $e \in \mathbb{R}^{d_e}$ denote the embedding from the original text description and pretrained language encoder, where $d_e$ is the text embedding dimension.

After passing the linear projection layer $Proj: \mathbb{R}^{d_e} \rightarrow \mathbb{R}^{d_p}$, the text embedding $e$ is prepended to $\mathbf{m}_d$. The final input to DMC is $\mathbf{x}=[Proj(e); \mathbf{m}_d]$, where $e$ is treated as the first token of the motion.
This allows the model to condition the entire refinement process on the text information while keeping the motion features structured.
Our $\mathrm{DMC}_\theta$ is a Transformer encoder with its parameter $\theta$, whose goal is to map noisy motion $\mathbf{m}_d$ to refined motion $\mathbf{m}_r$ when text embedding $e$ is given. The refinement process can be summarized as below:%
\begin{equation}
    \mathbf{m}_{r} = \mathrm{DMC}_\theta(\mathbf{x}) = \mathrm{DMC}_\theta([Proj(e); \mathbf{m}_d])
    \label{eq:dmc}
\end{equation}

\subsection{Self-supervised Distortion Refinement}
Motion generation models tend to produce motions with physically implausible artifacts (i.e., foot skating, floating, clipping, and ground penetration)~\cite{yuan2023physdiff, Han_2025_WACV}. In addition, they often exhibit over-smoothing along the temporal axis, which further blurs the details.
To mitigate these issues without relying on complex and costly physics modeling, we propose a framework for self-supervised distortion refinement. 

\vspace{1mm}
\noindent \textbf{Emulating Distortion for Self-supervision.    }
To prepare self-supervised learning, we first synthesize physically implausible motions by applying artificial distortions to ground-truth motions $\mathbf{m}_{\text{gt}}$ from the HumanML3D data~\cite{guo2022generating}. 
Specifically, we apply two types of distortion: (1) a random vertical offset $b$ applied along the Y-axis, and (2) a temporal Gaussian smoothing filter $\mathrm{g}(\cdot;\sigma)$ with a randomly sampled standard deviation $\sigma$. The vertical offset $b$ synthesizes contact-related artifacts such as foot floating (i.e., $b>0$) or ground penetration (i.e., $b<0$). The Gaussian smoothing filter introduces varying levels of temporal inconsistency or smoothing with varying values of $\sigma$.

By exposing DMC to such a spectrum of distorted inputs, it learns to effectively refine physically implausible motions during inference. The overall process is summarized below:%
\begin{equation}
\scalebox{0.91}{$
    \mathbf{m}_{d} = \mathrm{g}(\mathbf{m}_{\text{gt}} + [0, b, 0]; \sigma),
    \quad
    \parbox{4cm}{
        \raggedright
        $b \sim \text{Uniform}(-0.1,\ 0.1)$ \\
        $\sigma \sim \text{Uniform}(0.1,\ 4.0)$
    }
    \label{eq:distortion}
$}
\end{equation}%
Note that the range of $b$ and $\sigma$ is empirically selected to approximate the spectrum of artifacts that are commonly observed in pretrained text-to-motion generative models. 

\vspace{1mm}
\noindent \textbf{Strategies to Distortion Refinement Learning. }
We explore two strategies to train DMC based on triplets $(\mathbf{m}_d, e, \mathbf{m}_{\text{gt}})$: (1) a fast adversarial refinement and (2) an iterative denoising-based refinement.

\noindent \textbf{(1) WGAN-based DMC.   } First strategy is to train our DMC by using the Wasserstein GAN with Gradient Penalty (WGAN-GP)~\cite{hook2017wgangp} framework. Here, $\mathrm{DMC}_\theta(\cdot)$ acts as a generator $G$ that refines $\mathbf{m}_d$ into $\mathbf{m}_r = G([Proj(e); \mathbf{m}_d])$. For a discriminator $D$, we choose a Vision Transformer-based model~\cite{dosovitskiy2021vit} that takes both the sentence embedding and refined motion as input by concatenating them in the token space, and provides adversarial feedback to guide the generator toward more plausible motions. $G$ is trained using a combination of adversarial loss and reconstruction loss, as defined below:%
\begin{align}
    \mathcal{L}_G &= -D(\mathbf{m}_r) + \lambda \big\| \mathbf{m}_r - \mathbf{m}_{\text{gt}} \big\|_2^2,
    \label{eq:gen_loss}
    \\
    \mathbf{m}_r &= G\big([Proj(e); \mathbf{m}_d]\big) = \mathrm{DMC}_\theta\big([Proj(e); \mathbf{m}_d]\big), \nonumber
\end{align}%
The first term of $\mathcal{L}_G$ leads $G$ to generate motions that fool the discriminator, while the second term penalizes discrepancies between $\mathbf{m}_r$ and $\mathbf{m}_{\text{gt}}$. The hyperparameter is initialized to $\lambda=5$, and it gradually decays during training. $D$ is trained using the gradient penalty objective~\cite{hook2017wgangp}, as shown below:%
\begin{equation}
    \mathcal{L}_D = D(\mathbf{m}_{r})  -  D(\mathbf{m}_{\text{gt}}) + \gamma 
 ( \| \nabla D(\hat{\mathbf{m}}) \|_2 - 1 )^2,
    \label{eq:disc_loss}
\end{equation}%
where $\hat{\mathbf{m}}$ is a linear interpolation between $\mathbf{m}_{\text{gt}}$ and $\mathbf{m}_{r}$ used for computing the gradient penalty. 
Here, we empirically choose $\gamma=10$ for the gradient penalty hyperparameter. 

\vspace{1mm}
\noindent \textbf{(2) Denoiser-based DMC.   } Inspired by denoising diffusion probabilistic models (DDPMs)~\cite{ho2020ddpm}, our second strategy treats DMC as iterative denoiser, which repeatedly refines $\mathbf{m}_d$ to $\mathbf{m}_r$.
However, unlike DDPM which denoises Gaussian random noise into a data sample, we instead focus on learning to denoise distortions in $\mathbf{m}_d$.
To guide the denoising process at each training step, we sample a random timestep $t \in \{1, \dots, T\}$, and construct an intermediate noisy input $\mathbf{m}_t$ by mixing the clean and distort motions as below:%
\begin{equation}
    \mathbf{m}_t = \frac{T - t}{T} \; \mathbf{m}_{\text{gt}} + \frac{t}{T} \; \mathbf{m}_{d}, \quad t \in \{1, \dots, T\}.
    \label{eq:xt}
\end{equation}

In this case, $\mathrm{DMC}_\theta$ is trained to predict the residual to reverse this interpolation, i.e., $\mathbf{m}_{t} - \mathbf{m}_{t-1}$.
The denoising loss function $\mathcal{L}_N$ is as below, with $t$ as a positional encoding:%
\begin{equation}
    \mathcal{L}_N = \big\| \mathrm{DMC}_\theta\big([Proj(e); \mathbf{m}_t]; t\big) - (\mathbf{m}_{t} - \mathbf{m}_{t-1}) \big\|_2^2.
    \label{eq:denoise_loss}
\end{equation}

During inference, this DMC refines the motion over the denoising steps $\hat{T}$, which can be freely adjusted by users based on the severity of physical artifacts. The below equations demonstrate the inference phases:%
\begin{equation}
    \mathbf{m}_{\hat{T}} = \mathbf{m}_d,
    \;
    \mathbf{m}_{t-1} = \mathbf{m}_{t} - \mathrm{DMC}_\theta\big([Proj(e); \mathbf{m}_t]; t\big)
\end{equation}

We evaluate the model performance with various lengths $\hat{T} \in \{1, 10, 20, 30, ..., 100\}$, and find out that properly chosen $\hat{T}$ can handle the degree of physical implausibility of $\mathbf{m}_d$. It can yield better refinement, with empirically selected values of $\hat{T}=50$ for T2M and T2M-GPT, and $\hat{T}=10$ for MoMask.

\afterpage{
\begin{table*}[t]
\caption{Comparison study on the HumanML3D data. For each metric, the average performance over 20 runs with 95\% confidence intervals is reported. \textbf{Bold} indicates the best result within each baseline (T2M, T2M-GPT, MoMask), while \underline{underline} denotes the second best.}
\label{tab:comparison-baselines}
\centering
\resizebox{\textwidth}{!}{
\begin{tabular}{l c ccc c ccccc}
\toprule
\multirow{2}{*}{Method} & \multirow{2}{*}{FID$\downarrow$} & \multicolumn{3}{c}{R-Precision$\uparrow$} & \multirow{2}{*}{MPJPE$\downarrow$} & \multirow{2}{*}{Skate$\rightarrow$} & \multirow{2}{*}{Float (m)$\rightarrow$} & \multirow{2}{*}{Penetrate (m)$\downarrow$} & \multirow{2}{*}{Clip (m)$\downarrow$} \\
\cmidrule(lr){3-5}
& & Top 1 & Top 2 & Top 3 \\
\midrule
Real motion & -- & 0.5206$^{\pm .0039}$ & 0.7094$^{\pm .0024}$ & 0.8005$^{\pm .0024}$ & 19.6273$^{\pm .0000}$ & 0.0733$^{\pm .0002}$ & 2.6430$^{\pm .0083}$ & 0.0000$^{\pm .0000}$ & 0.0000$^{\pm .0000}$ \\
\midrule
T2M~\cite{guo2022generating} & 1.0037$^{\pm .0253}$ & 0.4567$^{\pm .0032}$ & 0.6368$^{\pm .0027}$ & 0.7395$^{\pm .0029}$ & \textbf{22.2421}$^{\pm .0320}$ & \textbf{0.0783}$^{\pm .0004}$ & 1.1918$^{\pm .0268}$ & \underline{0.0296}$^{\pm .0008}$ & 0.0381$^{\pm .0005}$ \\
\; + Supv-DMC & \underline{0.6876}$^{\pm .0192}$ & \textbf{0.4648}$^{\pm .0032}$ & \textbf{0.6465}$^{\pm .0032}$ & \textbf{0.7483}$^{\pm .0031}$ & 22.6289$^{\pm .0322}$ & 0.1020$^{\pm .0004}$ & 1.0715$^{\pm .0263}$ & 0.3014$^{\pm .0022}$ & \underline{0.0377}$^{\pm .0004}$ \\
\; + WGAN-DMC & \textbf{0.5747}$^{\pm .0177}$ & \underline{0.4634}$^{\pm .0030}$ & \underline{0.6462}$^{\pm .0031}$ & \underline{0.7468}$^{\pm .0034}$ & 22.5057$^{\pm .0332}$ & 0.0497$^{\pm .0003}$ & \underline{1.2084}$^{\pm .0266}$ & 0.0661$^{\pm .0018}$ & \textbf{0.0374}$^{\pm .0005}$ \\
\; + Denoise-DMC & 0.8780$^{\pm .0231}$ & 0.4619$^{\pm .0031}$ & 0.6423$^{\pm .0030}$ & 0.7438$^{\pm .0030}$ & \underline{22.3626}$^{\pm .0321}$ & \underline{0.0604}$^{\pm .0003}$ & \textbf{1.2139}$^{\pm .0275}$ & \textbf{0.0170}$^{\pm .0006}$ & 0.0380$^{\pm .0005}$ \\
\midrule
T2M-GPT~\cite{zhang2023t2mgptgeneratinghumanmotion} & 0.2099$^{\pm .0040}$ & 0.4622$^{\pm .0032}$ & 0.6469$^{\pm .0028}$ & 0.7459$^{\pm .0022}$ & \underline{22.6658}$^{\pm .0291}$ & \textbf{0.0557}$^{\pm .0004}$ & \underline{1.7048}$^{\pm .0360}$ & 0.2370$^{\pm .0053}$ & 0.0386$^{\pm .0004}$ \\
\; + Supv-DMC & \textbf{0.1648}$^{\pm .0043}$ & 0.4589$^{\pm .0032}$ & 0.6454$^{\pm .0031}$ & 0.7461$^{\pm .0022}$ & 22.7450$^{\pm .0322}$ & 0.0969$^{\pm .0006}$ & 1.4630$^{\pm .0349}$ & 0.8403$^{\pm .0069}$ & \underline{0.0380}$^{\pm .0005}$ \\
\; + WGAN-DMC & 0.2209$^{\pm .0076}$ & \textbf{0.4644}$^{\pm .0031}$ & \textbf{0.6531}$^{\pm .0027}$ & \textbf{0.7520}$^{\pm .0024}$ & \textbf{22.3207}$^{\pm .0310}$ & \underline{0.0552}$^{\pm .0004}$ & 1.7030$^{\pm .0356}$ & \textbf{0.1147}$^{\pm .0038}$ & \textbf{0.0376}$^{\pm .0005}$ \\
\; + Denoise-DMC & \underline{0.1822}$^{\pm .0031}$ & \underline{0.4625}$^{\pm .0026}$ & \underline{0.6471}$^{\pm .0026}$ & \underline{0.7464}$^{\pm .0025}$ & 22.8175$^{\pm .0294}$ & 0.0460$^{\pm .0004}$ & \textbf{2.1343}$^{\pm .0371}$ & \underline{0.2113}$^{\pm .0048}$ & 0.0385$^{\pm .0004}$ \\
\midrule
MoMask~\cite{guo2023momask} & \underline{0.0457}$^{\pm .0019}$ & \underline{0.5199}$^{\pm .0024}$ & \underline{0.7114}$^{\pm .0022}$ & \textbf{0.8056}$^{\pm .0020}$ & 23.1478$^{\pm .0379}$ & \underline{0.0711}$^{\pm .0003}$ & \underline{2.4286}$^{\pm .0236}$ & \underline{0.0812}$^{\pm .0018}$ & 0.0381$^{\pm .0003}$ \\
\; + Supv-DMC & \textbf{0.0424}$^{\pm .0018}$ & 0.5179$^{\pm .0027}$ & 0.7100$^{\pm .0021}$ & 0.8043$^{\pm .0017}$ & \underline{23.0884}$^{\pm .0372}$ & 0.1142$^{\pm .0005}$ & 1.9998$^{\pm .0225}$ & 0.3772$^{\pm .0035}$ & \textbf{0.0377}$^{\pm .0004}$ \\
\; + WGAN-DMC & 0.0741$^{\pm .0030}$ & 0.5135$^{\pm .0025}$ & 0.7056$^{\pm .0023}$ & 0.8015$^{\pm .0017}$ & \textbf{22.9977}$^{\pm .0375}$ & \textbf{0.0715}$^{\pm .0003}$ & 2.2822$^{\pm .0249}$ & 0.1050$^{\pm .0035}$ & \underline{0.0378}$^{\pm .0003}$ \\
\; + Denoise-DMC & 0.0459$^{\pm .0020}$ & \textbf{0.5200}$^{\pm .0023}$ & \textbf{0.7115}$^{\pm .0022}$ & \underline{0.8055}$^{\pm .0018}$ & 23.1758$^{\pm .0379}$ & 0.0641$^{\pm .0003}$ & \textbf{2.4500}$^{\pm .0234}$ & \textbf{0.0544}$^{\pm .0017}$ & 0.0381$^{\pm .0003}$ \\
\bottomrule
\end{tabular}
}
\vspace{-5mm}
\end{table*}
}

\section{Experiments}

\subsection{Dataset and Evaluation Metrics}
\textbf{Dataset.  }
We use HumanML3D~\cite{guo2022generating} as the main benchmark to evaluate the physical plausibility and semantic consistency of generated and refined motions. The dataset contains 14,616 3D motion sequences paired with 44,970 natural language descriptions, spanning a diverse vocabulary of 5,371 words. Each motion is annotated with at least three distinct captions, and the average duration is 7.1 seconds (ranging from 2 to 10 seconds). To ensure temporal consistency, all motions are uniformly resampled at 20 FPS.

\vspace{1mm}
\noindent \textbf{Evaluation Metrics.  }
To assess the effectiveness of DMC in correcting physical artifacts while preserving semantic alignment, we adopt five standard evaluation metrics commonly used in prior motion generation studies~\cite{guo2022generating, guo2023momask, Han_2025_WACV}, covering both physical plausibility and text-motion consistency: (1) \textbf{FID (Fr\'echet Inception Distance)}: Measures the distributional similarity between refined motions and real motion features. Lower scores indicate better perceptual quality and overall realism. (2) \textbf{R-Precision}: Evaluates the alignment between the generated motion embeddings and text embeddings based on retrieval accuracy. The metric measures whether the correct textual description is included within the top-3 nearest candidates, assessing how well the motions refined by DMC are semantically aligned with the input textual description. (3) \textbf{Foot Skating Ratio}: Following the definition in GMD~\cite{karunratanakul2023gmd}, we compute the ratio in which the foot is in contact with the ground but moves more than 2.5cm, indicating temporal inconsistency. A ratio closer to that of the ground-truth motion indicates more natural grounding, as it reflects realistic foot contact behavior. (4) \textbf{Foot Floating and Ground Penetration}: Following ReinDiffuse~\cite{Han_2025_WACV}, we measure the distances between the lowest joint and the ground plane in frames if the contact occurs. Positive values indicate floating, and negative values indicate penetration. These values reflect how well DMC corrects vertical misalignment artifacts. (5) \textbf{Foot Clipping}: Following the definition in ReinDiffuse~\cite{Han_2025_WACV}, we measure the severity of foot interpenetration when the distance between both feet drops below a threshold (5cm). Lower values indicate more physically coherent corrections by DMC. 



\vspace{1mm}
\subsection{Quantitative Results}
We quantitatively evaluate DMC,
after it is applied
to three baseline text-to-motion models: T2M~\cite{guo2022generating}, T2M-GPT~\cite{zhang2023t2mgptgeneratinghumanmotion}, and MoMask~\cite{guo2023momask}. Table~\ref{tab:comparison-baselines} shows results on the metrics above. Note that each experiment is repeated 20 times, and we report the mean values with standard deviations. 

\vspace{1mm}
\noindent \textbf{Comparison of Training Strategies with Baselines.  }
As shown in Table~\ref{tab:comparison-baselines}, DMC consistently improves motion quality across all baselines by enhancing either perceptual realism or physical plausibility.
We first compare three training strategies for DMC: (1) \textbf{supervised training}, (2) \textbf{WGAN-based training}, and (3) \textbf{denoising-based training} to examine their respective impacts on perceptual quality, semantic consistency, and physical plausibility.

\vspace{1mm}
\noindent \textbf{Supervised Training.  }
We suggest the supervised variant of DMC as a comparison baseline, which learns to refine the distorted motion with a simple mean squared loss function.
To be specific, it is trained to regress the vertical bias $b$ and the smoothing scale $\sigma$, while simultaneously refining the motion.
The supervised variant of DMC shows moderate improvements in perceptual quality and semantic alignment. For instance, FID improves by 31.49\% on T2M, 21.49\% on T2M-GPT, and 7.22\% on MoMask. It also enhances R-Precision on T2M and T2M-GPT.
Notably, even for a high-quality baseline like MoMask, which already achieves strong FID and R-Precision scores, the supervised strategy yields measurable improvement in perceptual quality.
However, this strategy exhibits limited effectiveness in correcting physical artifacts. In some cases, it even degrades physical plausibility. For instance, penetration errors increase from 0.0296 to 0.3014 on T2M and from 0.2370 to 0.8403 on T2M-GPT. This suggests that while the supervised model is effective at learning global perceptual features, it lacks robustness in resolving fine-grained physical inconsistencies due to the inherent bias in its supervision signal. Given these limitations, we chose not to adopt this strategy.

\vspace{1mm}
\noindent \textbf{WGAN-based Training.   } 
The WGAN-based strategy excels at improving perceptual realism and semantic alignment. It achieves the best FID reduction on T2M, improving the score by 42.74\%, and records the highest R-Precision on T2M-GPT. Notably, DMC raises the R-Precision score of T2M beyond that of the original T2M-GPT baseline, indicating improvement in semantic alignment. Furthermore, it significantly reduces contact-related artifacts, with a 51.6\% decrease in ground penetration on T2M-GPT and a 2.59\% reduction in clipping on T2M. However, it is relatively less effective than denoising-based training in resolving subtle artifacts like floating or fine-grained contact errors. While it offers only modest gains in physical plausibility, its primary strength lies in globally aligning motion appearance with textual semantics. Additionally, since it operates in one refinement step, WGAN-based DMC is computationally efficient and particularly suitable for scenarios requiring fast refinement and improved semantic consistency.

\vspace{1mm}
\noindent \textbf{Denoising-based Training.   }
The denoising-based DMC demonstrates strong performance in refining physical realism. It reduces ground penetration by 42.57\% on T2M, 10.84\% on T2M-GPT, and 33.0\% on MoMask. Notably, when applied to MoMask, the R-Precision improves to 0.52, the best all methods, and both the floating and penetration errors are significantly reduced. This highlights that denoising-based refinement can further enhance physical plausibility, even for already high-quality motions. 
This superior correction capability stems from its iterative refinement process, where motion is gradually denoised in a step-wise manner. Such progressive correction enables the model to resolve subtle physical inconsistencies that are difficult to address in single-step refinement. While improvements in FID are moderate compared to WGAN-based training, it still maintains competitive perceptual quality and semantic alignment. Due to the increased computational cost from multi-step inference, this strategy is best suited for applications demanding high-fidelity motion with minimal physical artifacts. Notably, for high-quality baselines like MoMask, fewer denoising steps are needed, making the trade-off between refinement quality and speed more favorable.

\afterpage{
\begin{figure*}[t]
  \centering
  \includegraphics[width=0.99\linewidth]{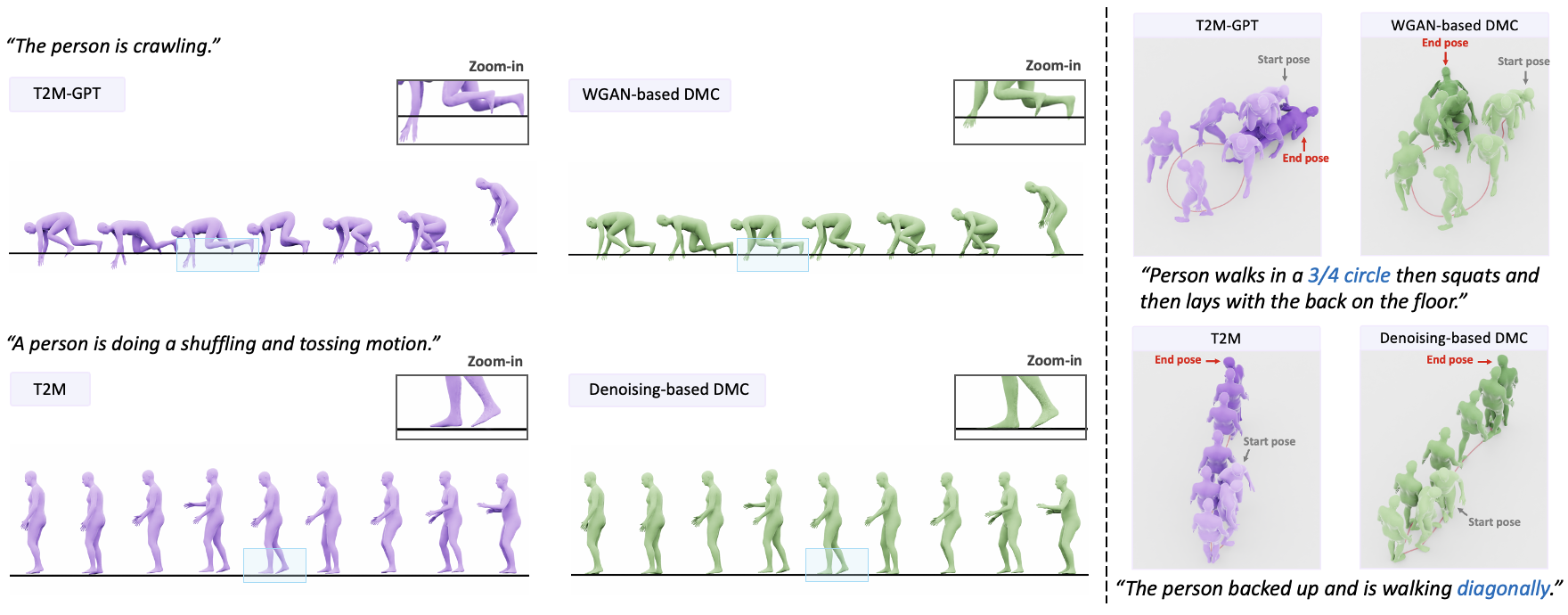}
  \captionof{figure}{Qualitative comparisons on the HumanML3D dataset. We visualize motion sequences generated by T2M-GPT (top) and T2M (bottom), and their corresponding refined results using the proposed DMC. \textbf{(1) Left:} Our DMC significantly improves physical plausibility, effectively correcting not only severe ground penetration (top) but also subtle foot floating (bottom), as highlighted in the zoom-in views. \textbf{(2) Right:} In addition, we show motion samples before and after refinement using DMC, where color gradients indicate temporal progression from bright to dark shades, demonstrating more physically plausible and consistent trajectories.}
  \label{fig:qualitative_all}
  \vspace{-1ex}
\end{figure*}
}

\vspace{1mm}
\noindent \textbf{Comparison and Recommendation.   }
As discussed earlier, WGAN-based and denoising-based training strategies exhibit complementary strengths. The WGAN-based DMC efficiently refines artifact-prone baselines with semantic fidelity, requiring only 0.4 ms per sample. In contrast, the denoising-based DMC achieves more accurate correction of fine-grained artifacts through iterative refinement, with processing times of 335.18 ms at timestep 100, 167.98 ms for T2M and T2M-GPT at timestep 50, and 33.16 ms for MoMask at timestep 10. Accordingly, WGAN-based DMC is preferable when fast inference and semantic consistency are required, while denoising-based DMC is recommended for precise physical correction.


\afterpage{
\begin{table*}[t]
  \centering
    \caption{Ablation study of the text embedding $e$ under the WGAN training strategy. We evaluate the effect of incorporating CLIP text embeddings across three baselines. While FID and R-Precision show comparable results with and without $e$, physical plausibility consistently improves with text guidance.}
    \label{tab:ablation-text-wgan}
    \centering
    \resizebox{\textwidth}{!}{
    \begin{tabular}{ll cccc cccc}
      \toprule
      \multirow{2}{*}{Method} & \multirow{2}{*}{Strategy} & \multirow{2}{*}{FID$\downarrow$} & \multicolumn{3}{c}{R-Precision$\uparrow$} & \multirow{2}{*}{Skate$\rightarrow$} & \multirow{2}{*}{Float (m)$\rightarrow$} & \multirow{2}{*}{Penetrate (m)$\downarrow$} & \multirow{2}{*}{Clip (m)$\downarrow$} \\
      \cmidrule(lr){4-6}
      & & & Top 1 & Top 2 & Top 3 \\
      \midrule
      Real motion & -- & -- & 0.5206$^{\pm .0039}$ & 0.7094$^{\pm .0024}$ & 0.8005$^{\pm .0024}$ & 0.0733$^{\pm .0002}$ & 2.6430$^{\pm .0083}$ & 0.0000$^{\pm .0000}$ & 0.0000$^{\pm .0000}$ \\
      \midrule
      T2M~\cite{guo2022generating} & w/o $e$ & 0.5860$^{\pm .0181}$ & 0.4624$^{\pm .0032}$ & 0.6459$^{\pm .0032}$ & 0.7463$^{\pm .0035}$ & \textbf{0.0561}$^{\pm .0003}$ & 1.1650$^{\pm .0263}$ & 0.0902$^{\pm .0019}$ & 0.0375$^{\pm .0005}$ \\
                                 \; + WGAN-DMC & w/ $e$ & \textbf{0.5747}$^{\pm .0177}$ & \textbf{0.4634}$^{\pm .0030}$ & \textbf{0.6462}$^{\pm .0031}$ & \textbf{0.7468}$^{\pm .0034}$ & 0.0497$^{\pm .0003}$ & \textbf{1.2084}$^{\pm .0266}$ & \textbf{0.0661}$^{\pm .0018}$ & \textbf{0.0374}$^{\pm .0005}$ \\
                                 
      \midrule
      T2M-GPT~\cite{zhang2023t2mgptgeneratinghumanmotion} & w/o $e$ & \textbf{0.2063}$^{\pm .0072}$ & 0.4639$^{\pm .0034}$ & 0.6526$^{\pm .0027}$ & 0.7516$^{\pm .0024}$ & \textbf{0.0608}$^{\pm .0005}$ & 1.5969$^{\pm .0357}$ & 0.1581$^{\pm .0042}$ & 0.0379$^{\pm .0006}$ \\
                                     \; + WGAN-DMC  & w/ $e$  & 0.2209$^{\pm .0076}$ & \textbf{0.4644}$^{\pm .0031}$ & \textbf{0.6531}$^{\pm .0027}$ & \textbf{0.7520}$^{\pm .0024}$ & 0.0552$^{\pm .0004}$ & \textbf{1.7030}$^{\pm .0356}$ & \textbf{0.1147}$^{\pm .0038}$ & \textbf{0.0376}$^{\pm .0005}$ \\
      \midrule
      MoMask~\cite{guo2023momask} & w/o $e$ & 0.0745$^{\pm .0030}$ & 0.5125$^{\pm .0026}$ & 0.7049$^{\pm .0024}$ & 0.8010$^{\pm .0018}$ & 0.0775$^{\pm .0003}$ & 2.2329$^{\pm .0248}$ & 0.1284$^{\pm .0036}$ & \textbf{0.0377}$^{\pm .0003}$ \\
      
                              \; + WGAN-DMC & w/ $e$ & \textbf{0.0741}$^{\pm .0030}$ & \textbf{0.5135}$^{\pm .0025}$ & \textbf{0.7056}$^{\pm .0023}$ & \textbf{0.8015}$^{\pm .0017}$ & \textbf{0.0715}$^{\pm .0003}$ & \textbf{2.2822}$^{\pm .0249}$ & \textbf{0.1050}$^{\pm .0035}$ & 0.0378$^{\pm .0003}$ \\
      \bottomrule
    \end{tabular}
    }
    \vspace{-4ex}
\end{table*}

}

\subsection{Qualitative Result}
To further verify the effectiveness of DMC, we visualize motion sequences generated from T2M~\cite{guo2022generating} and T2M-GPT~\cite{zhang2023t2mgptgeneratinghumanmotion} before and after applying refinement with DMC.
The visualization results are shown in Figure~\ref{fig:qualitative_all}. 

Figure~\ref{fig:qualitative_all} (right) demonstrates the semantic refinement capability of DMC. In the top row, the input description includes a complex term, “3/4 circle.” The baseline motion generated by T2M-GPT completes a full circular path, overshooting the intended trajectory. In contrast, the motion refined by the WGAN-based DMC more closely follows the 3/4 circle instruction, correcting the semantic discrepancy. The bottom row shows motion generated by T2M, where the character walks diagonally but fails to reflect the intended diagonal direction in the description. After applying DMC trained with a denoising-based strategy, the trajectory becomes more aligned with the given textual description, showing a diagonal walk.

In Figure~\ref{fig:qualitative_all} (left), we demonstrate the impact of DMC on improving physical plausibility. The top row illustrates a crawling motion generated by T2M-GPT, where severe hand-ground penetration is clearly visible. After applying WGAN-based DMC, results become more realistic with better physically grounded contact. In the bottom row, the result by T2M has a shuffling and tossing motion, where subtle foot floating artifacts are present. Denoising-based DMC successfully refines these inconsistencies, yielding precise contact and better overall realism.
These qualitative results show that DMC is effective in both enhancing physical plausibility, across varying degrees of artifact severity, improving the semantic alignment of generated motions with the given textual descriptions.

\subsection{Ablation Study}
\noindent \textbf{Effect of Text Embeddings.    }
DMC is designed to correct physically implausible motion while preserving alignment with the input textual description. To examine the effect of textual guidance, we compare two settings under WGAN-based training: using CLIP embeddings~\cite{radford2021clip} and without any text embedding. Results across three base models are presented in Table~\ref{tab:ablation-text-wgan}.

Interestingly, the absence of text embeddings does not lead to substantial degradation in FID or R-Precision scores, indicating that DMC can perform perceptually coherent refinement even without explicit semantic input. However, we observe consistent improvements in physical plausibility metrics, particularly in foot floating and ground penetration, when CLIP embeddings are used. For example, in the case of T2M-GPT, ground penetration is reduced by 51.60\% with text embeddings, whereas the reduction drops to 33.29\% without them. 
Due to page limitation, detailed results are omitted, but similar trends were observed under denoising-based training.

These findings indicate that text embedding, while having limited influence on perceptual quality or high-level semantics, contributes meaningfully to the refinement process by encouraging motions that are more physically grounded. Rather than operating solely as a physical-level post-processing module, DMC functions as a language-guided refinement mechanism that leverages textual cues to enhance contact accuracy and physical plausibility.

\begin{table}[t]
  \centering
  \caption{Comparison of Denoising-based DMC performance under different distortion types used for self-supervised training. Vertical bias ($B$) induces vertical offset in motion, while Gaussian smoothing ($S$) applies temporal filtering to create overly smooth motions.}
  \label{tab:ablation-distortion-comparison}
  \resizebox{\linewidth}{!}{ 
  \begin{tabular}{l cc ccccc}
    \toprule
    Method & $B$ & $S$ & Skate$\rightarrow$ & Float (m)$\rightarrow$ & Penetrate (m)$\downarrow$ & Clip (m)$\downarrow$ \\
    \midrule
    Real motion & \ding{55} & \ding{55} & 0.0733 & 2.6430 & 0.0000 & 0.0000 \\
    \midrule
    \multirow{3}{*}{\shortstack[l]{T2M~\cite{guo2022generating} \\ + Denoise-DMC}}
      & \ding{51} & \ding{55} & 0.0955 & 1.1161 & 0.0962 & 0.0381 \\
      & \ding{55} & \ding{51} & \textbf{0.0747} & 1.1873 & 0.0330 & 0.0381 \\
      & \ding{51} & \ding{51} & 0.0604 & \textbf{1.2139} & \textbf{0.0170} & \textbf{0.0380} \\
    \midrule
    \multirow{3}{*}{\shortstack[l]{T2M-GPT~\cite{zhang2023t2mgptgeneratinghumanmotion} \\ + Denoise-DMC}}
      & \ding{51} & \ding{55} & \textbf{0.0751} & 1.6196 & 0.3360 & 0.0386 \\
      & \ding{55} & \ding{51} & 0.0539 & 1.7028 & 0.2435 & \textbf{0.0384} \\
      & \ding{51} & \ding{51} & 0.0460 & \textbf{2.1343} & \textbf{0.2113} & 0.0385 \\
    \midrule
    \multirow{3}{*}{\shortstack[l]{MoMask~\cite{guo2023momask} \\ + Denoise-DMC}} 
      & \ding{51} & \ding{55} & 0.0787 & 2.3601 & 0.0964 & 0.0381 \\
      & \ding{55} & \ding{51} & \textbf{0.0700} & 2.4243 & 0.0826 & 0.0380 \\
      & \ding{51} & \ding{51} & 0.0641 & \textbf{2.4500} & \textbf{0.0544} & \textbf{0.0381} \\
    \bottomrule
  \end{tabular}
  }
\vspace{-4ex}
\end{table}

\vspace{1mm}
\noindent \textbf{Effect of Distortion Types for Self-supervision.  }DMC learns to reconstruct physically plausible motion from the distorted inputs without relying on ground-truth supervision. The type of distortion applied during training will determine which artifacts the model learns to correct. To investigate this, we evaluate DMC under distortion configurations below:

\begin{itemize}
    \item \textbf{Vertical Bias Only}: Applies random vertical offsets to the entire motion sequence along the y-axis, resulting in artifacts such as foot floating or ground penetration.
    \item \textbf{Smoothing Only}: Applies Gaussian smoothing to temporal trajectories, leading to over-smoothed sequences that exhibit foot skating.
    \item \textbf{Vertical Bias + Smoothing}: Combines vertical misalignment with temporal smoothing to jointly simulate contact-related artifacts.
\end{itemize}

The results are shown in Table~\ref{tab:ablation-distortion-comparison}. Each distortion type induces distinct errors, and the model trained on that distortion shows improvement in the corresponding metric. 

Interestingly, applying both vertical bias and smoothing simultaneously leads to stronger overall improvements across all physical plausibility metrics. In particular, we observe that the gains from this combined distortion setting surpass the improvements achieved by each distortion alone. This suggests that exposing DMC to multiple types of artifacts during training is crucial for robust and generalizable refinement performance. Also, it shows that DMC benefits most when trained to correct a wide range of physical errors, rather than specializing in a single artifact type.

\section{CONCLUSION}
In this work, we introduce the \textbf{Distortion-aware Motion Calibrator (DMC)}, a lightweight and model-agnostic post-hoc refinement framework designed to enhance physical plausibility and better align generated motions with their intended textual semantics. 
Rather than modifying the base architecture or relying on heavy physical modeling, DMC corrects common artifacts (e.g., foot floating) through a self-supervised refinement process.

We propose two DMC variants that jointly improve semantic and physical alignment: (1) a WGAN-based model that excels in enhancing perceptual quality and semantic consistency, and (2) a denoising-based model that offers more fine-grained correction of contact-related artifacts. Extensive experiments on three representative baselines (T2M~\cite{guo2022generating}, T2M-GPT~\cite{zhang2023t2mgptgeneratinghumanmotion}, and MoMask~\cite{guo2023momask}) demonstrate that DMC consistently improves both physical plausibility and semantic alignment across varying motion qualities. 

Our current implementation only models limited distortions, leaving broader artifacts such as jittering or self-intersections unaddressed. Expanding the distortion set and embedding robot-specific physical constraints (e.g., mass, torque, and speed limits) would enhance generalizability and physical realism. Such extensions hold promise for improving text-to-motion retargeting in real-world humanoid scenarios.

By operating as a plug-and-play module that requires no re-training of the base model, DMC opens up new possibilities for scalable and efficient post-processing in human motion generation pipelines. We believe it can serve as a practical solution for integrating physically realistic motion into applications such as character animation, virtual agents, and robotics.








\bibliographystyle{ieeetr}
\bibliography{IEEEabrv, refs}

\end{document}